# Towards Smart Monitored AM: Open Source in-Situ Layer-wise 3D Printing Image Anomaly Detection Using Histograms of Oriented Gradients and a Physics-Based Rendering Engine

*Aliaksei Petsiuk[1] and Joshua M. Pearce[2*]*

1. Department of Electrical & Computer Engineering, Michigan Technological University, Houghton, MI, USA
2. Department of Electrical & Computer Engineering, Western University, ON, Canada

apetsiuk@mtu.edu, * joshua.pearce@uwo.ca

## Graphical abstract

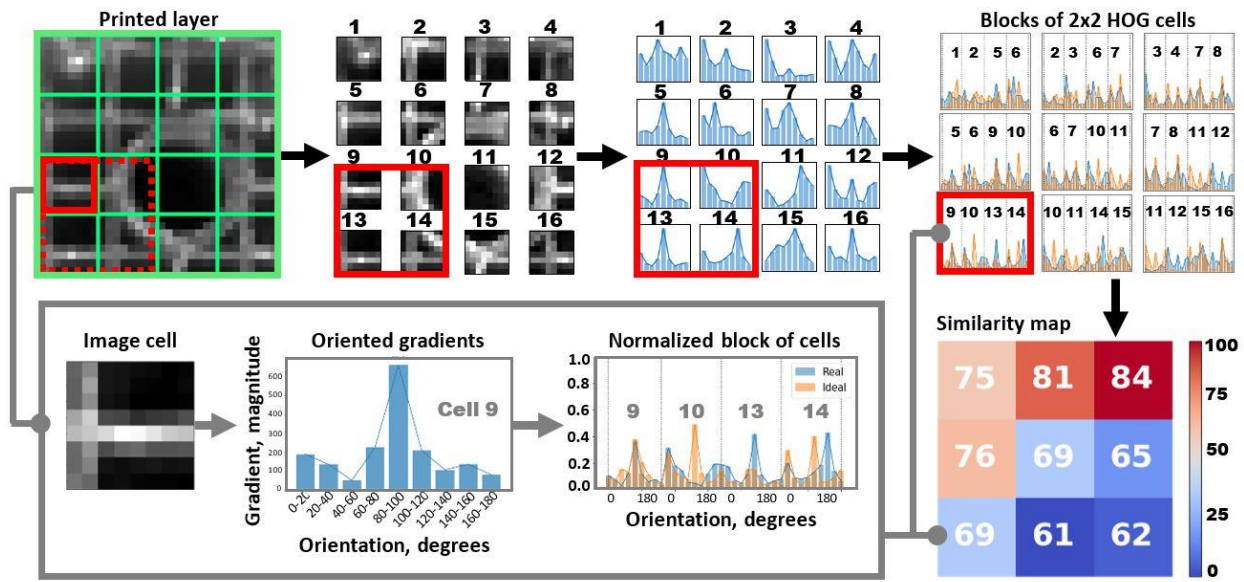

## Highlights

- In-situ layer-wise 3D printing anomaly detection is based on reference images.
- G-code-based synthetic references are created in a physics rendering engine.
- The system analyzes the similarity of local histograms of oriented gradients.
- Failure detection resolution is 5-10% of the entire observation area.
- The method allows noticing critical errors in the early stages of their occurrence.

## Abstract

This study presents an open source method for detecting 3D printing anomalies by comparing images of printed layers from a stationary monocular camera with G-code-based reference images of an ideal process generated with Blender, a physics rendering engine. Recognition of visual deviations was accomplished by analyzing the similarity of histograms of oriented gradients (HOG) of local image areas. The developed technique requires preliminary modeling of the working environment to achieve the best match for orientation, color rendering, lighting, and other parameters of the printed part. The output of the algorithm is a level of mismatch between printed and synthetic reference layers. Twelve similarity and distance measures were implemented and compared for their effectiveness at detecting 3D printing errors on six different representative failure types (local infill defects, presence of a foreign body in the layer, spaghetti problem, separation and shift of the printing part from the working surface, defects in thin walls, and layer shift) and their control error-free print images. The results show that although Kendall's tau, Jaccard, and Sorensen similarities are the most sensitive, Pearson's r, Spearman's rho, cosine, and Dice similarities produce the more reliable results. This open source method allows the program to notice critical errors in the early stages of their occurrence and either pause manufacturing processes for further investigation by an operator or in the future AI-controlled automatic error correction. The implementation of this novel method does not require preliminary data for training, and the greatest efficiency can be achieved with the mass production of parts by either additive or subtractive manufacturing of the same geometric shape. It can be concluded this open source method is a promising means of enabling smart distributed recycling for additive manufacturing using complex feedstocks as well as other challenging manufacturing environments.

**1. Introduction**

Over the past decades, additive manufacturing (AM) has become a widespread technology that has found application in various fields of science and technology. AM allows the fabrication of high-performance components with complex geometries and continues to attract research interest. Extrusion-based 3D printing, democratized with the open source release of the self-replicating rapid prototyper (RepRap) [1-3], dominates the technology arena due to its low cost [4], availability of components, and a wide variety of printing materials [5,6] including waste plastics [7-10]. Despite its affordability and relative ease of use, however, this technology is not free from fabrication failures, which reduces economic impact [11,12], environmental merits [13], and limits the prospects for industrialization [14,15].

According to a recent comprehensive state-of-the-art review of monitoring techniques for material extrusion AM [16], the number of publications in the field of anomaly analysis grows steadily as this is a major impediment to widespread deployment. The vast majority of research has been conducted in the fused filament fabrication (FFF) domain [16]. This phenomenon can be explained by the fact that FFF technology dominates the 3D printing market for printers in use [17].

Analysis of extrusion-based AM processes can consist of examining parameters such as temperature [18,19], vibration [20,21], acoustic emissions [22,23], electrical characteristics [24,25], and others [26-29]. The main source of information, however, remains 2D and 3D image data obtained from single or multiple camera systems [16]. Since 3D printed parts are mostly fabricated in layers, most of the developed failure detection methods analyze manufacturing processes after a certain number of layers have been printed. Nuchitprasitchai et al. [30], Johnson et al. [31], and Hurd [32] proposed the concepts of failure analysis based on comparison with Standard Tessellation Language (STL) files. Jeong et al. [33] and Wasserfall et al. [34] employed information obtained from G-code files of printing parts. Ceruti et al. [35] utilized data from computer-aided design (CAD) files. Researchers also use comparison with reference data [36,37] or ideal printing processes [38,39]. Malik et al. [40] presented a 3D reconstruction-based scanning method for real-time monitoring of AM processes.

Having a way to automatically detect critical errors will significantly reduce material waste and time spent on failed prints. In order to reach this goal, this study reports on a developed monocular system for the analysis of plastic FFF processes that monitors contour deviations and infill distortions for each layer. This work expands on previous developments of the authors [41] by using an open-source physics rendering engine to generate G-code-based synthetic reference images for each printing stage. With certain rendering parameters, a synthetic image can represent a real captured layer under ideal printing conditions. It is hypothesized that further comparative texture analysis based on image processing techniques can reveal the location and the degree of structural deviations. To this end a material extrusion-based 3D printer was monitored with a stationary monocular camera. Synthetic reference images for the setup were created with open-source and free Blender. Images were compared based on the similarity degree of the feature descriptors, represented by histograms of oriented gradients. Twelve similarity and distance measures were implemented and compared for their effectiveness at detecting 3D printing errors on six different representative failure types (local infill defects, presence of a foreign body in the layer, spaghetti problem, separation and shift of the printing part from the working surface, defects in thin walls, and layer shift) and their control error-free print images. The sensitivities of the measures are quantified and the results are discussed in the context of creating artificial intelligence (AI)-guided smart additive and subtractive manufacturing devices.

## 2. Method

### 2.1. Experimental apparatus

For experimental tests, an open source delta-style FFF-based 3D printer [42] was used, which represents a derivative of the self-replicating rapid prototype (RepRap) printer [1-3]. The device operates in Cartesian coordinates under the control of a RepRap Arduino Mega Polulu Shield (RAMPS) system [43]. It has a cylindrical volume of Ø240x250 mm and an extruder with a 0.4 mm nozzle diameter. The main feedstock material is 1.75 mm polylactic acid (PLA) plastic filament.

A stationary monocular camera is mounted on a tripod near the printing bed at an angle of ~45 degrees. The camera is based on a 2-megapixel 1/2.9-inch Sony IMX322 CMOS sensor [44] and has manual focus and aperture control. As shown in Figure 1, the four white dots on the printing surface are visual markers for the camera position and orientation determination. The markers indicate the active 90x90 mm observation area with the origin in the center of the square.

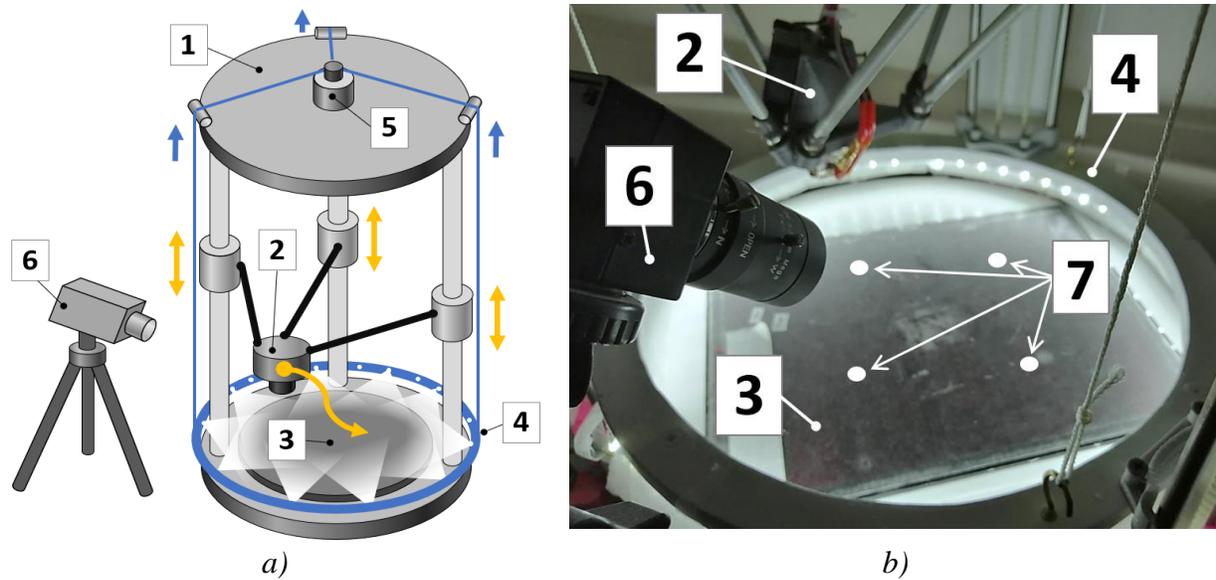

**Figure 1. Experimental apparatus:** *a)* 3D printer schematic, *b)* printing area. 1 – 3D printer, 2 – extruder, 3 – printing bed, 4 – movable circular lighting platform, 5 – lighting platform drive system, 6 – camera, 7 – visual markers.

A movable circular lighting platform [41], controlled through G-code, is located above the working surface. The light frame consists of 56 light-emitting diodes with a glowing temperature of 6000K and a total power of 18 watts. The stepper motor driving the mechanical structure is located on top of the printer and is connected to the RAMPS controller as an additional extruder.

**Table 1. Interlayer G-code commands**

| G-code command | Description |
| --- | --- |
| M400 | Wait for moves to finish |
| G91 | Switch to relative coordinates |
| G1 E-20 F1000 | Retract the filament 10 mm before lifting the nozzle |
| G1 Z80 | Move the nozzle 80 mm up |
| G1 X20 Y20 | Move the nozzle 20 mm aside |
| T1 | Set the active extruder to 1 (lighting platform) |
| G1 E-0.25 F600 | Move the lighting platform one layer height up |
| M400 | Wait for moves to finish |
| M42 P57 S200 | Indicator ON (optional) |
| **Create layer snapshot** | |
| G1 X-20 Y-20 | Move the nozzle 20 mm back |
| G1 Z-80 | Move the nozzle 80 mm down |
| G4 P500 | Wait 500ms for the nozzle vibration to stabilize |
| T0 | Set the active extruder to 0 (extruder nozzle) |
| G90 | Switch to absolute coordinates |
| M42 P57 S0 | Indicator OFF (optional) |

Table 1 shows a set of G-code commands that are placed after the printing instructions for each layer. This pauses the fabrication process, moves the extruder out of the video surveillance area, and raises the lighting platform to a height equal to the thickness of the printed layer. Thus, the active print area is evenly lit around the perimeter, regardless of the current layer and the working level of the extruder nozzle. This allows capturing 2D images of each completed layer with uniform illumination and applying unified image processing techniques to each image frame.

## 2.2. Creation of synthetic reference images

Synthetic reference images represent the ideal 3D printed model fabricated in optimal conditions. The open-source and free Blender [45] software was used to create images for each layer during the printing process. It is a multifunctional software environment for 3D graphics. The set of Blender tools includes 3D modeling, lighting and animation control, texture editing, and photorealistic rendering. There is also a Python scripting interface for customizing and automating the entire production pipeline.

Figure 2 depicts a virtual model of the key parts of the Delta printer in use.

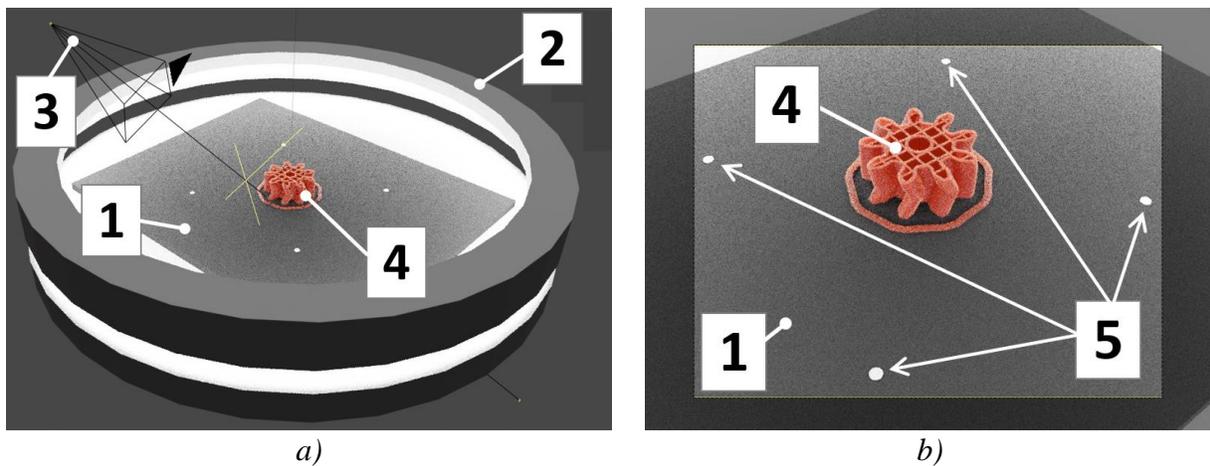

*a)*          *b)*

**Figure 2. Virtual workspace:** *a)* **main elements of the Delta printer modeled in Blender,** *b)* **virtual camera view area. 1 – printing bed, 2 – movable lighting platform, 3 – camera, 4 – rendered G-code, 5 – visual markers.**

Previous research has repeatedly shown that Blender can be used as a reliable and flexible physics simulating environment for solving scientific and engineering problems. Kent [46] utilized Blender to visualize astronomical data, Gschwandtner et al. [47] and Romulo Fernandes et al. [48] performed range sensor testing and radar simulations, respectively. Flaischlen and Wehinger [49] performed particle-resolved computational fluid dynamics modelling for chemical industry, Ilba [50] estimated solar irradiation on buildings, Rohe [51] created an optical test simulator, and, finally, Reitmann et al. [52] developed an add-on to generate semantically labeled depth-sensing data in Blender.

To create realistic images of the ideal printing process, the main components of the Delta printer have been modeled while maintaining relative proportions (Figure 2). Based on the authors' experiments and experience of professional 3D computer graphics communities [53,54], a shader graph was developed for the procedural generation of realistic plastic textures (Figure 3).

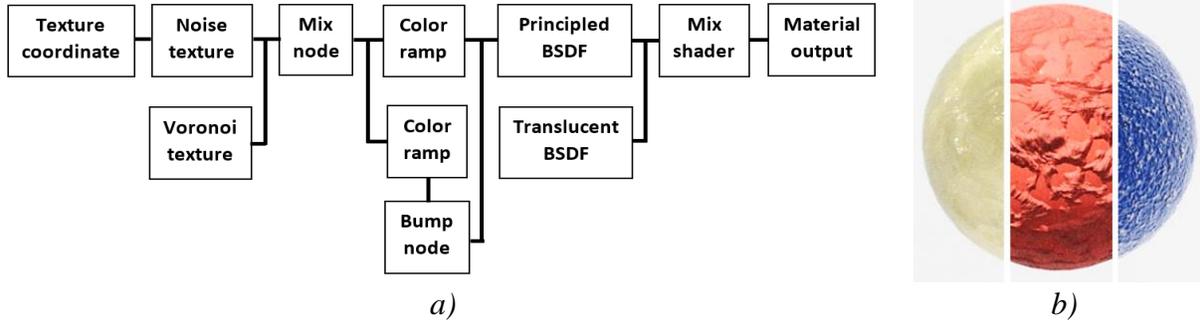

*a)* *b)*

**Figure 3. Shader graph for procedural texture generation:** *a)* **shader nodes,** *b)* **procedural texture samples**

The main nodes are the Principled and Translucent Bidirectional Scattering Distribution Functions (BSDFs). The Principled BSDF shader includes multiple material properties (roughness, reflection, transmission, sheen, etc.) as layers to create a wide variety of materials, and the Translucent BSDF adds Lambertian diffuse transmission [55]. The texture shaders, in turn, add natural surface irregularities. Changing the parameters of the nodes allows maximizing similarity with the real printed parts. The given graph (Figure 3, a) was used to visualize the photorealistic textures of the printed parts. In addition, a single material node with emissive characteristics was used to model the lighting frame.

To print a specific product, it is necessary to have its representation in the STL format, which describes the 3D object as a list of unit normals and vertices of its tessellated surface. During the process of slicing, the STL file is converted into G-code—a set of step-by-step trajectory coordinates for the printer extruder. Several G-code exporters [56-59] were used as references in this work.

An open source software toolchain has the G-code of the printing part to be loaded into the Blender programming interface and parsed layer by layer, where the extruder path is converted into a set of curves with an adjustable thickness parameter and preset material settings [60]. Therefore, each cross-section of the object can be represented as a G-code-based extruder path and an STL-based mask of the filled regions (Figure 4).

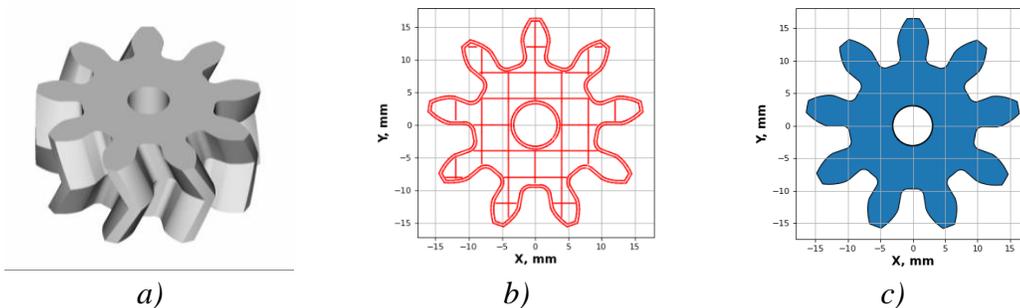

*a)* *b)* *c)*

**Figure 4. Printing object:** *a)* **STL file of the whole part,** *b)* **G-code of a layer cross-section,** *c)* **STL-based layer cross-section mask**

In the programmed 3D printing animation, a new printing layer is added with each consecutive frame, and the lighting platform is raised to the corresponding height until the "virtual print" is complete. Each frame is rendered with Blender Cycles [61], a physics-based path tracer, and saved as a separate image—a "quality standard" for comparison with the actual camera image of the printed layer.

## 2.3. Comparison of the printed layer with the reference image

The camera is positioned at an angle to the working area, which, however, makes it possible to visually rotate the active print surface (Figure 5) using a perspective projection (1) [62,63].

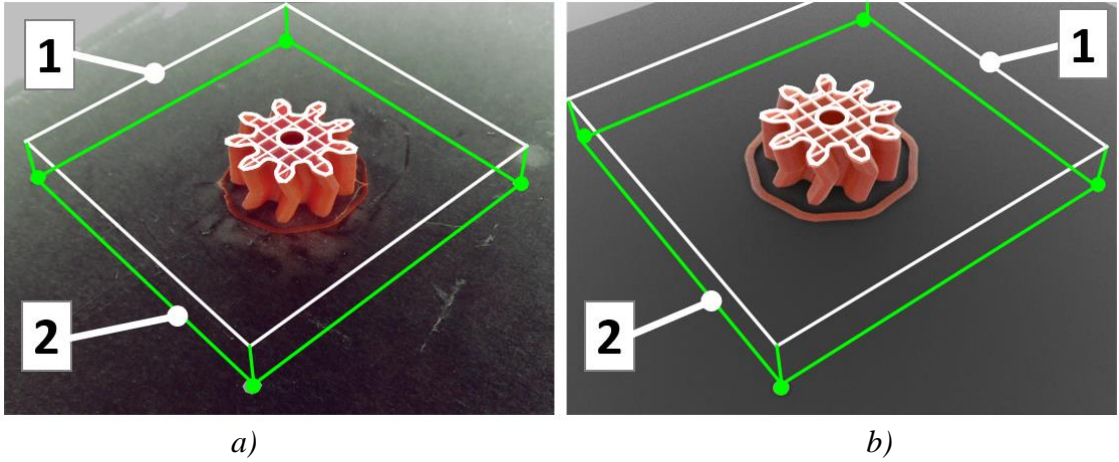

*a)*          *b)*

**Figure 5. Spatial position of the active printing area: *a)* real image, *b)* rendered image.
1 – active printing plane, 2 – print surface plane.**

Thus, regardless of the actual position and orientation of the camera, the virtual top views are used to analyze the AM process as if the camera was mounted directly above the printing bed (Figure 6).

$$\begin{bmatrix} tx' \\ ty' \\ t \end{bmatrix} = M \begin{bmatrix} x_p \\ y_p \\ 1 \end{bmatrix} = \begin{bmatrix} m_{11} & m_{12} & m_{13} \\ m_{21} & m_{22} & m_{23} \\ m_{31} & m_{32} & m_{33} \end{bmatrix} \cdot \begin{bmatrix} x_p \\ y_p \\ 1 \end{bmatrix} \quad (1)$$

Where $[x_p \quad y_p \quad 1]^T$ is the active area of the printed layer, $M$ is a projective transformation matrix, and $[tx' \quad ty' \quad t]^T$ is the virtual top view.

Each pixel of the virtual top view can be calculated based on the following equation (2):

$$(x', y') = \left( \frac{m_{11}x_p + m_{12}y_p + m_{13}}{m_{31}x_p + m_{32}y_p + m_{33}}, \frac{m_{21}x_p + m_{22}y_p + m_{23}}{m_{31}x_p + m_{32}y_p + m_{33}} \right) \quad (2)$$

It should be noted, however, that the video surveillance area is shifting upwards by the corresponding height with the printing of each new layer, so the unwrapped top view will remain orthogonal to the optical axis of the virtual top camera. Thus, after calculating the vertical shift, the 3D coordinates of the active printing plane $[X \quad Y \quad Z \quad 1]^T$ are projected onto the image frame to define the 2D boundaries $[x_p \quad y_p \quad 1]^T$ for unwrapping (3):

$$\begin{bmatrix} x_p \\ y_p \\ 1 \end{bmatrix} = K \begin{bmatrix} 1 & 0 & 0 & 0 \\ 0 & 1 & 0 & 0 \\ 0 & 0 & 1 & 0 \end{bmatrix} \begin{bmatrix} R_{3x3} & t_{3x1} \\ 0_{1x3} & 1 \end{bmatrix} \begin{bmatrix} X \\ Y \\ Z \\ 1 \end{bmatrix} \quad (3)$$

Where $[x_p \quad y_p \quad 1]^T$ is the active area projection onto the image plane, $K$ is the intrinsic camera parameters obtained during calibration, $R$ is the rotation matrix, $t$ is the translation vector, and $[X \quad Y \quad Z \quad 1]^T$ is the 3D coordinates of the active printing area.

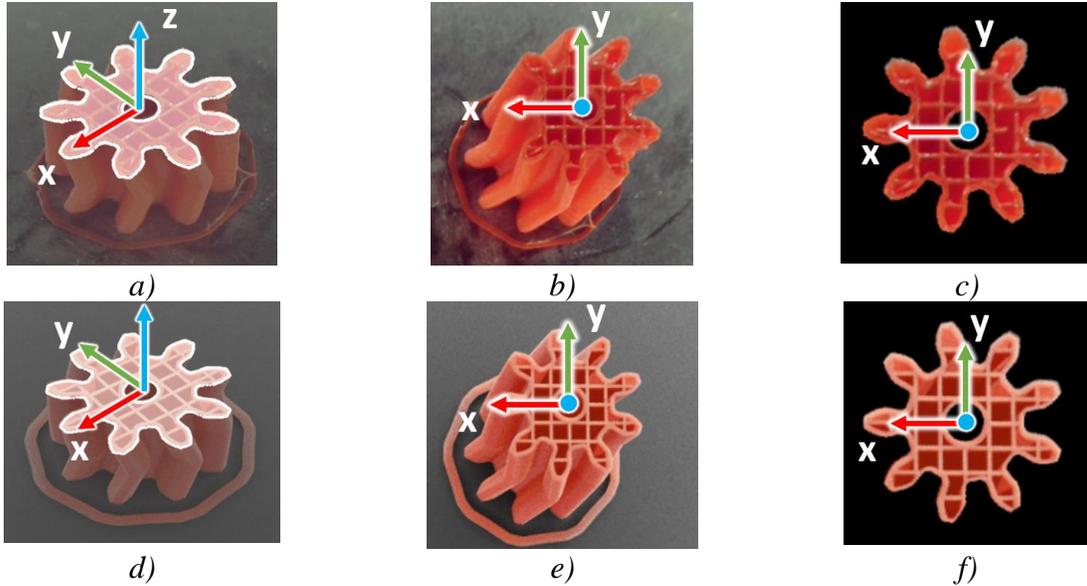

**Figure 6. Virtual top view:** *a)* **camera frame,** *b)* **unwrapped virtual top view,** *c)* **masked printing area,** *d)* **Blender scene frame,** *e)* **unwrapped synthetic image,** *f)* **masked rendered region**

Thus, knowing the camera location and the G-code coordinates, it is possible to rotate the printing area perpendicular to the camera axis, maintaining its origin in the center of the image (Figure 7).

After virtual rotation of the active printing plane, the real image is compared with the reference "ideal" one to analyze its texture and detect any possible defects inside the printed region. Image comparison is based on the similarity degree of the feature descriptors, represented by Histograms of Oriented Gradients (HOG) [64]. The descriptor analyzes local image areas, determines the orientation of the shaded gradients, and expresses this information as a histogram of direction channels.

HOG-based image analysis is widely applied in areas such as pattern recognition, template matching, and similarity determination. For example, Firuzi et al. [65] employed HOG features to recognize defects in electrical transformers, Malik et al. [66] presented a HOG-based landscape similarity analysis, Banerji et al. [67] enhanced HOG features with Fisher Model to extract geo-localization information from large-scale image datasets. Akila and Pavithra [68] developed an object detection algorithm based on scale invariant HOG descriptors, Joshi et al. [69] developed a sign language recognition system, and Supeng et al. [70] presented a HOG-based template matching algorithm.

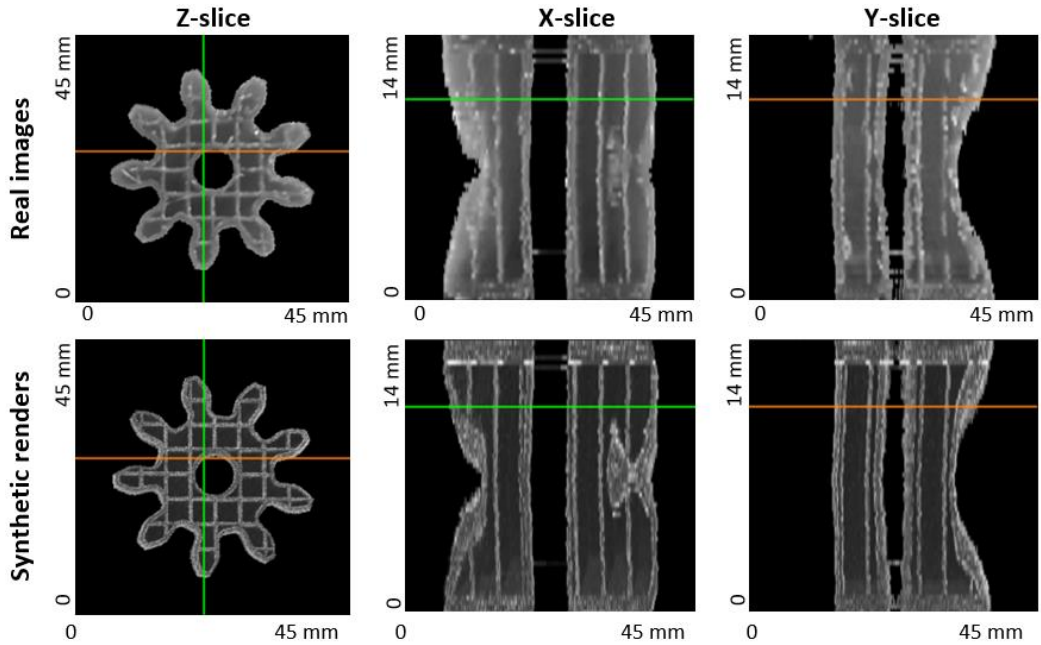

**Figure 7. Consecutive set of unwrapped layers combined into a volumetric view**

Figures 8 and 9 illustrate the detection of the dominant gradient orientation in local image areas and possible characteristic infill patterns with their feature descriptors, respectively. This allows capturing basic geometric structures by detecting the directions of contrasting edges.

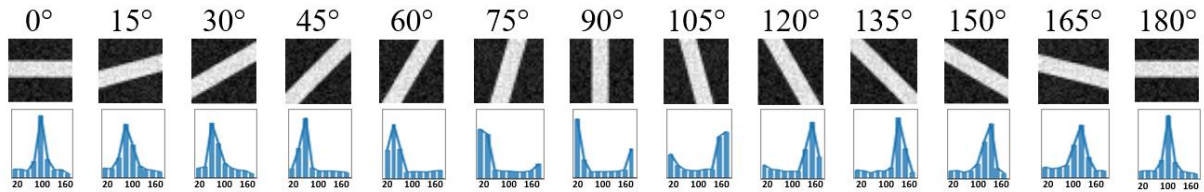

**Figure 8. Detection of the dominant gradient orientation in local image areas (top) using histograms of oriented gradients (bottom)**

As can be seen in Figure 9, changing the direction of light and shadow does not affect the determination of the dominant gradient orientation. The presence of noise, however, lowers the contrast, thereby limiting the capabilities of this method.

Figure 10 shows the stages of comparative image analysis. The image of the printed layer is divided into small sections, square cells, each of which is converted into a feature vector, represented by a nine-channel histogram of oriented gradients, ranging from 0° to 180° with 20-degree intervals. The feature vectors are then combined into normalized 2x2 blocks in such a way that each feature vector simultaneously contributes to several adjacent blocks, which increases the robustness of texture analysis. The same procedure is carried out for the reference synthetic image, after which the obtained histograms are compared. The result of this comparison is expressed in the form of a similarity map, where each section of the original camera image is assigned a numerical value of the degree of proximity to the "ideal" printing process. These values are then color coded to indicate ideal and non-ideal 3D printing.

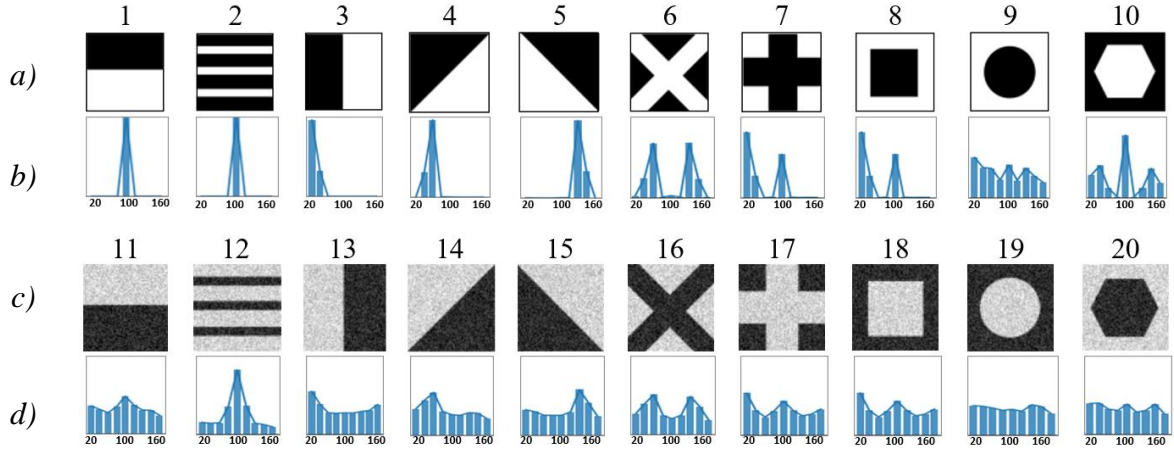

**Figure 9. Characteristic infill patterns (*a, c*) and their feature descriptors (*b, d*)**

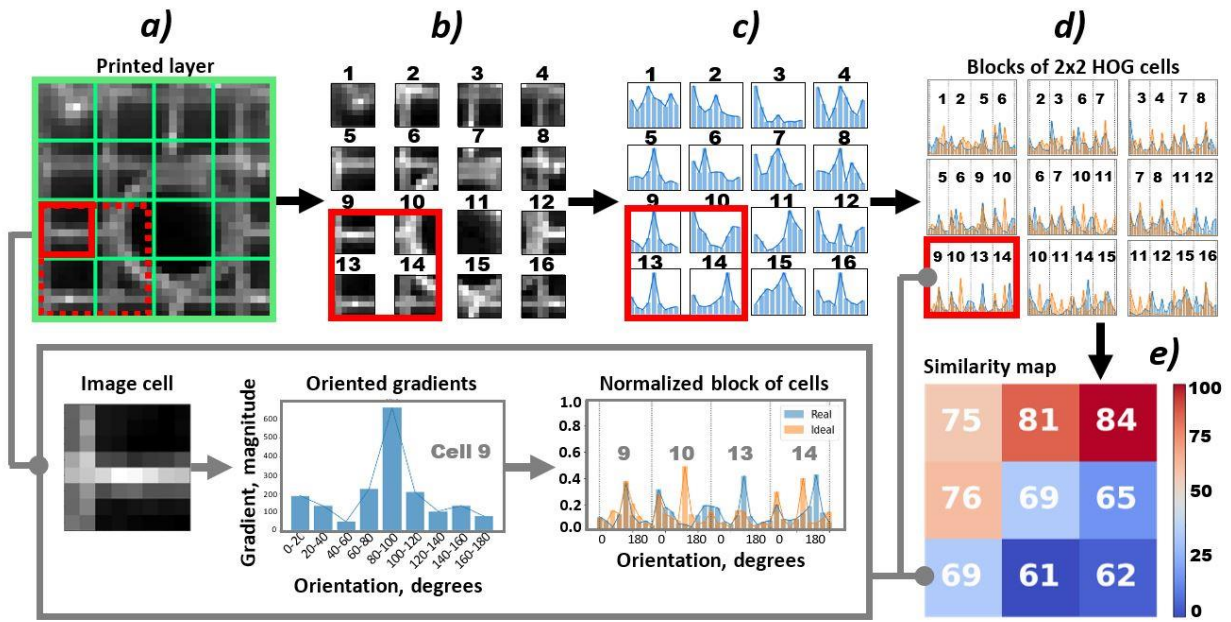

**Figure 10. Stages of comparative image analysis: *a)* splitting the original camera image into local areas, *b)* separate regions of the source image, *c)* converting image areas to feature vectors, *d)* comparison of normalized feature vectors of the original and reference images, *e)* resulting similarity map**

Thus, every unit section of the real image stores unique information from the corresponding area of the printed layer as a collection of features $\boldsymbol{p} = [p_1, p_2, \ldots, p_k]$. A single printed layer, in this way, can be represented as a tensor $\boldsymbol{F}$ with dimensions $\boldsymbol{N \times M \times k}$, where $\boldsymbol{N \times M}$ is the total number of image blocks, and $\boldsymbol{k}$ is the number of orientation channels of adjacent block cells. The tensor $\boldsymbol{F}$ is then transformed into an $\boldsymbol{N \times M}$ similarity matrix $\boldsymbol{D}$, each element $d_{ij}$ of which represents a similarity function $f(\mathbf{p}, \mathbf{q})$, where $\mathbf{q} = [q_1, q_2, \ldots, q_k]$ is a collection of features of the corresponding reference image block.

Comparison of histograms is done by determining the similarity $s(\mathbf{p}, \mathbf{q})$ between the corresponding vectors, $\mathbf{p}$ and $\mathbf{q}$. A number of similarity metrics were selected based on the comprehensive survey on similarity measures [71], a high-throughput X-ray diffraction pattern analysis [72], and in-depth quantitative analysis in the context of two real problems of image comparison and pattern location [73].

Hernandez–Rivera et al. [72] utilized 49 similarity measures to quantify similarities between Gaussian-based peak responses as a substitute for various characteristics in X-ray diffraction patterns. Research has not found universal metrics for all vector features. It was also found that the behavior of the metric response is not uniform for members of a given similarity family. It was determined, however, that the Clark metric yields a good balance between sensitivity and smooth changes.

Goshtasby et al. [73] found that Pearson correlation coefficient, Spearman's rho, Kendall's tau, Jaccard measure, $L_1$ norm, and squared $L_2$ norm overall perform better than other measures. Cosine similarity is also widely used in conjunction with HOG features in various pattern recognition tasks [70,74,75].

In this work, the twelve metrics shown in Table 2 were implemented.

**Table 2. Similarity and distance measures**

| Metric | Equation | Initial output range | Normalized output range |
|---|---|---|---|
| Cosine similarity | $s(\mathbf{p}, \mathbf{q}) = \dfrac{\sum p_i q_i}{\sqrt{\sum p_i^2 \sum q_i^2}}$ | [1, 0] | [1, 0] |
| Squared $L_2$ norm | $d(\mathbf{p}, \mathbf{q}) = \sum (p_i - q_i)^2$ | [0, 2] | [1, 0] |
| Pearson's r | $r(\mathbf{p}, \mathbf{q}) = \dfrac{\sum (p_i - \mu_p)(q_i - \mu_q)}{\sqrt{\sum (p_i - \mu_p)^2} \cdot \sqrt{\sum (q_i - \mu_q)^2}}$<br>Where $\mu_p = \frac{1}{n}\sum_{i=1}^{n} p_i$ and $\mu_q = \frac{1}{n}\sum_{i=1}^{n} q_i$ | [-1, 1] | [1, 0] |
| Spearman's rho | $\rho(\mathbf{p}, \mathbf{q}) = 1 - \dfrac{6 \sum d_i^2}{n(n^2 - 1)}$<br>Where $d_i$ is the difference between the two ranks of each observation, $n$ is the number of vector elements | [-1, 1] | [1, 0] |
| Kendall's tau | $\tau(\mathbf{p}, \mathbf{q}) = \dfrac{N_c - N_d}{n(n - 1)/2}$ | [-1, 1] | [1, 0] |

|  | Where $N_c$ and $N_d$ are the numbers of concordant and discordant pairs of vector elements, respectively |  |  |
|---|---|---|---|
| Jaccard | $s(\boldsymbol{p},\boldsymbol{q}) = \dfrac{\sum p_i q_i}{\sqrt{\sum(p_i^2 + q_i^2 - p_i q_i)}}$ | [1, 0] | [1, 0] |
| Dice | $s(\boldsymbol{p},\boldsymbol{q}) = \dfrac{2 \cdot \sum p_i q_i}{\sqrt{\sum(p_i^2 + q_i^2)}}$ | [1, 0] | [1, 0] |
| $L_1$ norm | $d(\boldsymbol{p},\boldsymbol{q}) = \sum |p_i - q_i|$ | [0, 2] | [1, 0] |
| Euclidean distance | $d(\boldsymbol{p},\boldsymbol{q}) = \sqrt{\sum (p_i - q_i)^2}$ | [0, $\sqrt{2}$] | [1, 0] |
| Hellinger distance | $d(\boldsymbol{p},\boldsymbol{q}) = \sqrt{2 \sum (p_i - q_i)^2}$ | [0, 2] | [1, 0] |
| Sorensen distance | $d(\boldsymbol{p},\boldsymbol{q}) = \dfrac{\sum |p_i - q_i|}{\sum (p_i + q_i)}$ | [0, 1] | [1, 0] |
| Clark distance | $d(\boldsymbol{p},\boldsymbol{q}) = \sqrt{\sum \left( \left|\dfrac{\|p_i - q_i\|}{p_i + q_i}\right|^2 \right)}$ | [0, $\sqrt{2}$] | [1, 0] |

Similarity metrics $s(\boldsymbol{p},\boldsymbol{q})$ are expected to satisfy the following properties [73,76]: limited range, $s \leq s_0$; reflexivity, $s(\boldsymbol{p},\boldsymbol{q}) = s_0$ if $\boldsymbol{p} = \boldsymbol{q}$; symmetry, $s(\boldsymbol{p},\boldsymbol{q}) = s(\boldsymbol{q},\boldsymbol{p})$, triangle inequality, $s(\boldsymbol{p},\boldsymbol{q}) \leq s(\boldsymbol{p},\boldsymbol{z}) + s(\boldsymbol{q},\boldsymbol{z})$, where $s_0$ is the largest measure between all possible vector inputs.

The input parameters $\boldsymbol{p}$ and $\boldsymbol{q}$ of the metrics shown in Table 2 are normalized vectors. The output range of similarity values is reduced to [1, 0], where "1" means the identity (overlap) of the input vectors, and "0" means a complete mismatch (no overlap). If a measure represents a distance between two inputs, $d(\boldsymbol{p},\boldsymbol{q})$, then the corresponding similarity is determined according to the following equation (4) [72]:

$$s(\boldsymbol{p},\boldsymbol{q}) = 1 - \frac{d(\boldsymbol{p},\boldsymbol{q})}{d^{max}(\boldsymbol{p},\boldsymbol{q})} \quad (4)$$

Where $d^{max}(\boldsymbol{p},\boldsymbol{q})$ is the absolute maximum possible distance between two input vectors for a particular measure. This method allows comparing all metrics, reducing them to the same scale.

Figure 11 shows the degrees of similarity of various input vectors in the forms of 9-bin histograms, expressed as a percentage of coincidence, reflecting the possible deviations of printed layer sections relative to the reference ones. To achieve maximum efficiency, it is necessary to obtain high similarity values for the first three cases, *a), b),* and *c),* and low similarity values for the latter cases, *d)* and *e).* This allows visualizing the efficacy of various similarity measures for specific analytical cases but does not reveal the capabilities of the metrics as applied to real-world problems. From Figure 11, however, it can be concluded that $L_1$ and $L_2^2$ norms, as well as Euclidean and Hellinger distances, produce results that are far from expected.

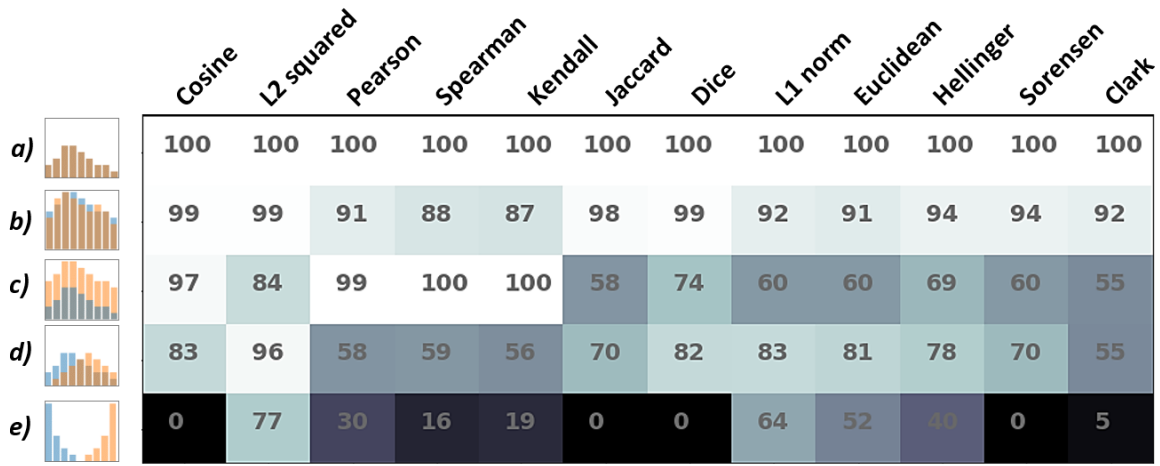

**Figure 11. Normalized similarity measures, expressed as a percentage of coincidence, for the following cases: *a)* complete match, *b)* similar histograms with small deviations, *c)* similar histograms with differences in levels (represents alike image areas with varying illumination parameters), *d)* histograms with significant shifts, *e)* non-overlapping histograms**

After the initial assessment of the effectiveness of the selected metrics, Pearson's r, Spearman's rho, Kendall's tau, as well as cosine, Jaccard, Dice, and Sorensen similarities were chosen to test the method for detecting printing errors.

## 3. Results

### 3.1 Test print modes for selecting optimal similarity measures

To select the optimal similarity metrics, real-life test images were used, reflecting the typical problems of 3D printing. Figure 12 shows the selected test images of erroneous layers for analyzing print mode abnormalities including: a) local infill defects, b) presence of a foreign body in the layer, c) spaghetti problem, d) separation and shift of the printing part from the working surface, e) defects in thin walls, and f) layer shift. A regular printed layer is provided for each failed case, which allows comparing the outputs for various printing regimes and calculating the discriminative power for the selected metrics. In addition to different types of defects, the selected parts have geometries of varying degrees of complexity.

Considering the camera parameters and the size of the working area, the scale of the captured images is 6.67 pixels per millimeter. The size of the minimum area of similarity analysis (2x2 block of 8-pixel cells) is therefore 4.8x4.8 mm, which lies in the range of 5-10% of the entire area of observation. This parameter can potentially be improved by using a high-definition camera.

In addition to continuous similarity for each local area of the image, an experimental failure threshold was also introduced. This is the main criterion for manufacturing defects, which allows varying sensitivity of the selected metrics and unequivocally segmenting the erroneous regions within the printed layers. Thus, an image area is considered defective if its match with the corresponding reference image is less than the chosen threshold.

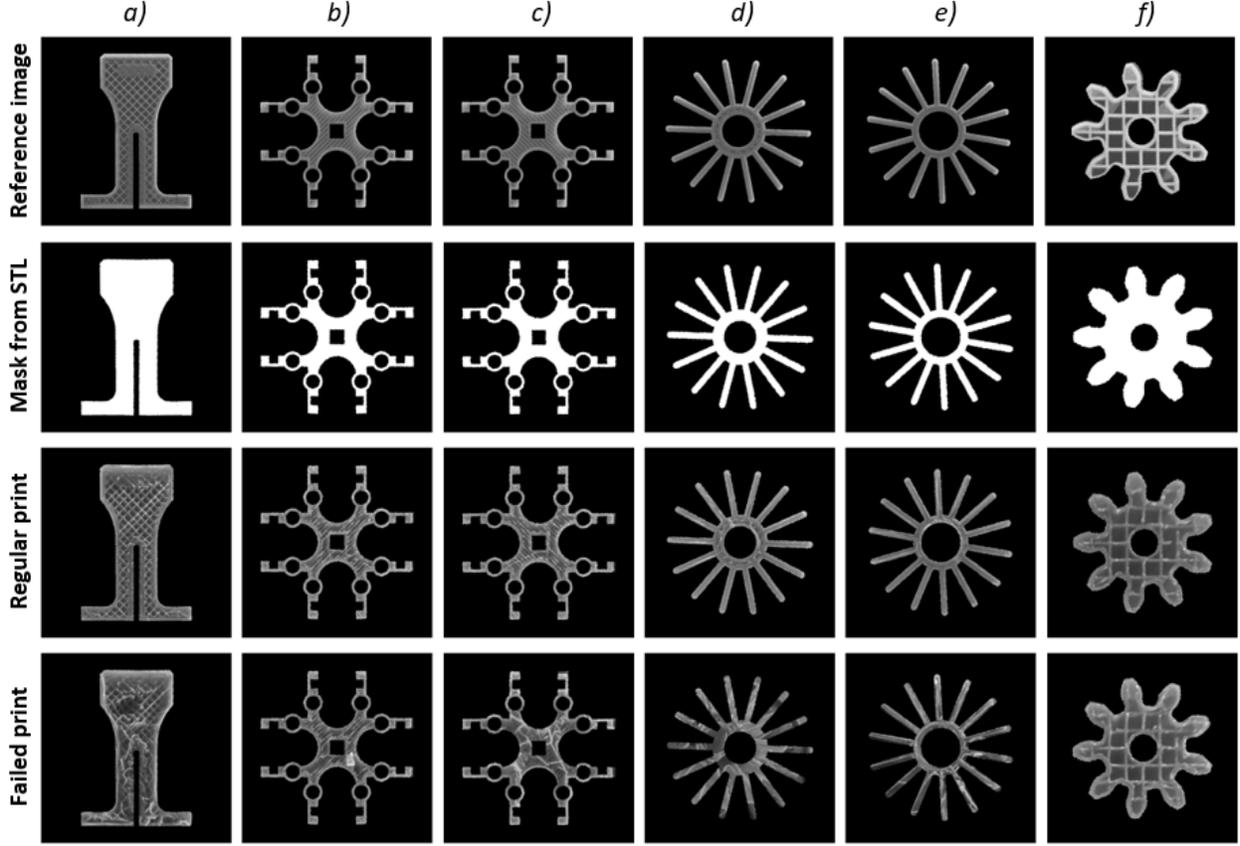

**Figure 12.** Test images of regular and failed printed layers: *a)* local infill defects, *b)* presence of a foreign body in the layer, *c)* spaghetti problem, *d)* separation and shift of the printing part from the working surface, *e)* defects in thin walls, *f)* layer shift

### 3.2 Comparing HOG-based similarity measures

Figure 13 depicts the similarities of test images in the form of heatmaps within the unit range. Each of the metrics can be used to analyze additive manufacturing processes. As can be seen from the heatmaps, Kendall's tau, Jaccard, and Sorensen similarities are the most sensitive, while Pearson's r, Spearman's rho, cosine, and Dice similarities produce more reliable results.

To determine the distinctive power of the selected metrics (Figure 14), an arbitrary 70% failure threshold ($T_S$) was applied to the calculated heatmaps ($H$). A layer region is considered normal if its similarity index is greater than or equal to $T_S$. Thus, the overall ratio of anomalous areas ($a_\%$) for each layer is calculated using the following expression (5):

$$a_\% = \frac{(H \leq T_S)}{S_P} \cdot 100\% \qquad (5)$$

Where $S_P$ is the area of the entire printed layer, $H$ is the layer similarity, and $T_S$ is the 70% failure threshold.

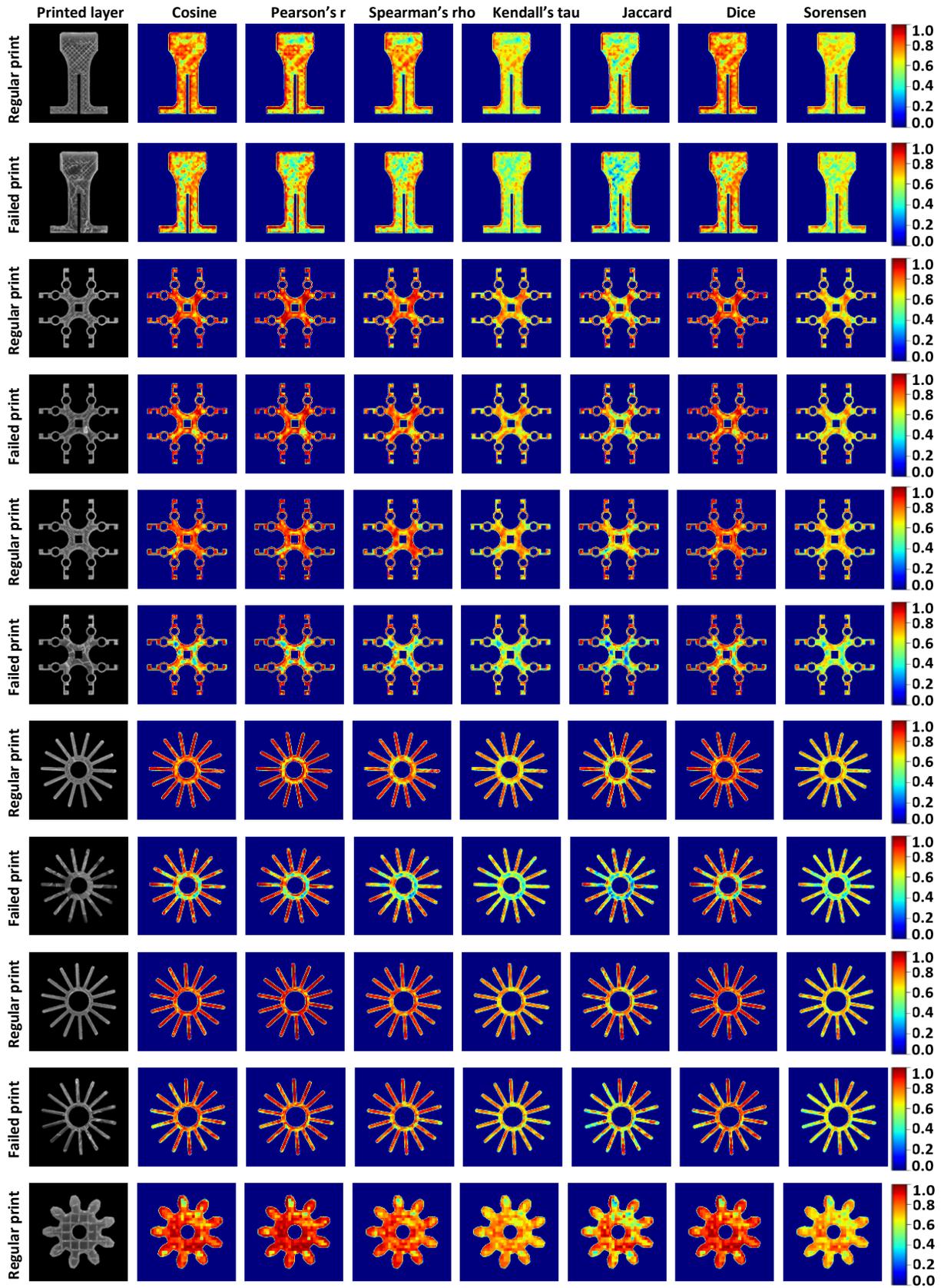

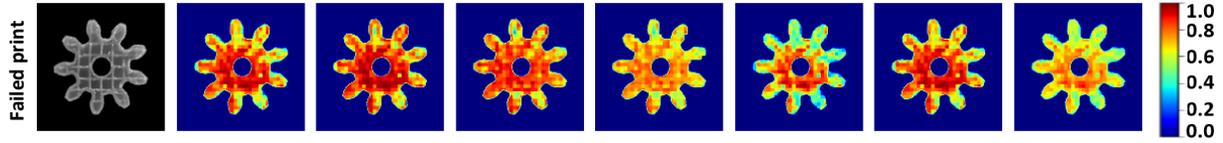

**Figure 13. Heat maps of the regular and failed prints for the example components**

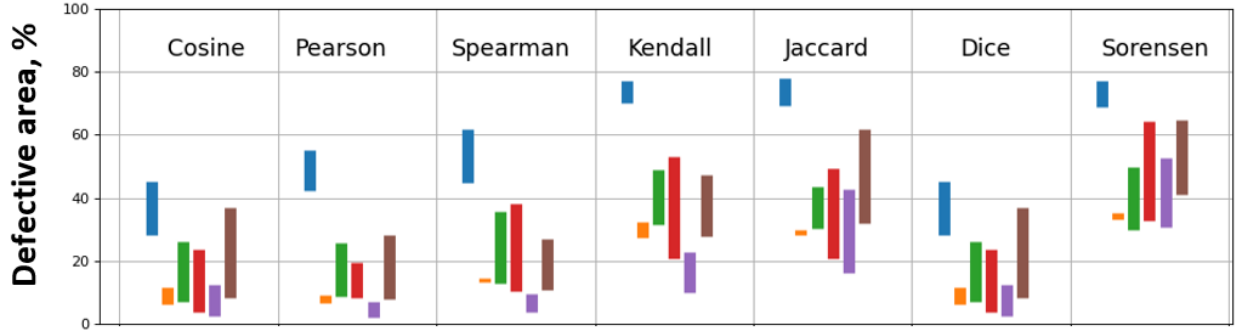

**Figure 14. Discriminative power of the selected metrics after applying the 70% failure threshold. Vertical color bars represent the difference between the regular (bottoms) and failed (tops) prints: local infill defects (blue), presence of a foreign body in the layer (orange), spaghetti problem (green), separation and shift of the printing part from the working surface (red), defects in thin walls (purple), layer shift (brown).**

Figure 15 shows an example of failure detection and segmentation for the case of cosine similarity.

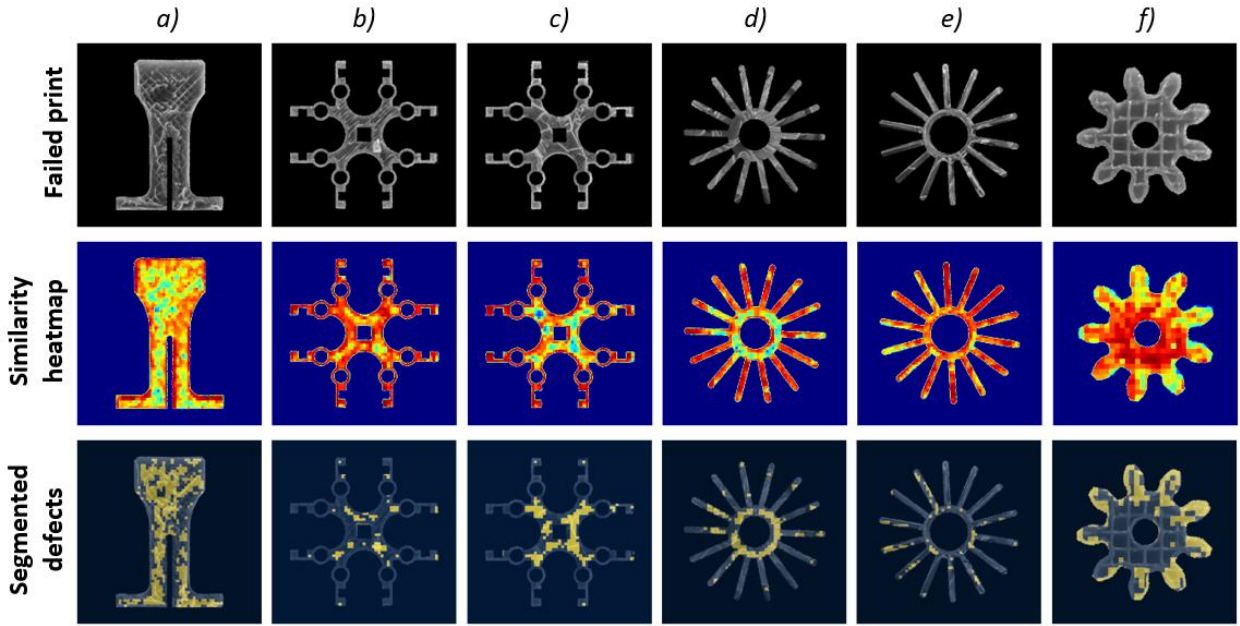

**Figure 15. Example of failure detection based on HOG features and cosine similarity:** *a)* **local infill defects,** *b)* **presence of a foreign body in the layer,** *c)* **spaghetti problem,** *d)* **separation and shift of the printing part from the working surface,** *e)* **defects in thin walls,** *f)* **layer shift**

## 4. Discussion

The proposed method makes it possible to analyze 3D printed parts for each layer and segment of anomalous regions with a size of 5 mm or more. The main limiting factor for the developed technique is the time-consuming preliminary rendering. Generating a single image with optimized render settings can take up to several minutes on an Intel Core i7 2.60 GHz system with a dedicated NVIDIA GeForce RTX 2060 GPU. The large size and complex geometric shape of the part increase visualization time. Since the image acquisition and analysis take less than 5 seconds, most of the time spent on in-situ monitoring can be attributed to the pause and nozzle retraction during the image capturing, which can consume more than 20 seconds of time.

There are several previously used approaches to analyze additive manufacturing processes based on information from 3D models (G-code, CAD, or STL) and reference images. Johnson et al. [31] proposed a method using an STL file to generate a binary cross-sectional image of a part corresponding to the layer being inspected. Malik et al. [40] implemented a method for real-time layer-by-layer monitoring of AM processes using G-code-based masking, 3D reconstruction, and augmented reality. Lyngby et al. [77] introduced a real-time vision system for non-nominal AM operation detection employing references from CAD models and color-based image segmentation. Wasserfall et al. [34] proposed a dual camera setup for in-situ layer-wise verification of 3D printed electronics based on known G-code geometry. Hurd et al. [32] presented an approach using a phone camera to validate layers quality by comparing AM process images with 2D images obtained from an STL file. Nuchitprasitchai et al. [30] developed single- and double-camera systems for detecting 5% deviations at every 50th layer utilizing shape and size data from STL files. Delli and Chang [78] proposed a binary 3D printing error detection technique based on support vector machine classification, where the quality check is being performed at critical printing stages based on available images of an ideal printing process. The methods listed above, however, do not employ a physics-based rendering engine and, therefore, do not allow using the rich set of image processing techniques.

The presented method is not limited to the use only in the field of 3D printing. This technique can be applied to compare parts produced by subtractive manufacturing and to reject defective printed circuit boards. The considered rendering system can be suitable for non-expert users to select parameters such as percentage and type of infill, wall thickness, and color of the material at the preprinting stage. Another application is 3D printing in a humanitarian context [79-83] where the work environment has unreliable power supplies or air quality issues such as pollution/debris in the processing environment. This also includes distributed recycling for additive manufacturing (DRAM) [7-10] with non-uniform and heavily contaminated filament. In addition, these methods can also be applied to pellet extrusion 3D printing [84-88] or fused particle fabrication (FPF) [89-93], where particles are direct 3D printed (sometimes as simply shredded waste), as such conditions are more likely to cause critical errors.

The described technique can be applied to FFF 3D printers of all sizes that can be imaged. The additional movable lighting platform used in this work is not necessary equipment since the fundamental factor is the coincidence of physical and virtual printing systems, regardless of lighting conditions. The original 3D model of the physical equipment can be easily customized to match the actual printing conditions for any type of 3D printer. In the future, the failure detection method can be improved by integrating a physics-based rendering system into the printing process to enable an AI monitored manufacturing process capable of real-time correction of anomalies from digital designs for both AM as well as subtractive manufacturing.

## 5. Conclusions

This study describes the conceptual capabilities of monocular layer-wise in-situ monitoring and analysis of additive manufacturing processes using projective transformations and image processing techniques. The developed method can be applied to material extrusion 3D printers of any size with a resolution of detecting anomalous regions of 5-10% of the overall observation area. The minimum area of similarity analysis can potentially be decreased by using a high-definition camera.

The results show that HOG-based similarity comparison does not introduce significant delays in the monitoring process. The described method, however, requires time resources when preparing the virtual environment and rendering the reference images. Using different similarity metrics and failure thresholds provides flexibility and allows for varying sensitivity in printing anomalies segmentation. Of the twelve similarity and distance measures implemented and compared for their effectiveness at detecting 3D printing errors, the results show that although Kendall's tau, Jaccard, and Sorensen similarities are the most sensitive, Pearson's r, Spearman's rho, cosine, and Dice similarities produce the more reliable results. The greatest efficiency of the given technique can be achieved with the mass production of parts of the same geometric shape as the Blender rendering only needs to occur once. Although this technique was tested here for additive manufacturing, it can be applied to compare parts produced by subtractive manufacturing and printed circuit boards. It can be concluded this method is a promising candidate for enabling smart manufacturing of all kinds in challenging environments such as those for humanitarian 3D printing or distributed recycling for additive manufacturing using complex waste feedstocks.

**CRediT authorship contribution statement**

**Aliaksei Petsiuk:** Conceptualization, Methodology, Software, Validation, Formal analysis, Investigation, Writing - original draft. **Joshua M. Pearce:** Conceptualization, Methodology, Formal analysis, Resources, Writing - original draft, review & editing, Supervision, Project administration, Funding acquisition.

**Declaration of Competing Interest**

The authors declare that they have no known competing financial interests or personal relationships that could have appeared to influence the work reported in this paper.

**Acknowledgments**

This work was supported by the Witte and Thompson Endowments.